\documentclass[10pt,journal,twocolumn]{IEEEtran}

\usepackage{graphicx}
\usepackage{caption}
\usepackage{stmaryrd}
\usepackage{amsfonts,latexsym,amssymb}
\usepackage[cmex10]{amsmath}
\usepackage{algorithm}
\usepackage{algorithmic}
\usepackage{multirow}

\usepackage{graphicx}
\usepackage{subfig}

\usepackage{hyperref}

\input{zgx.math}
\newtheorem{Proposition}{Proposition}
\newtheorem{remark}{Remark}

\usepackage{color}
\newcommand{\zgx}[1]{{#1}}

\begin{document}

\title{Group Component Analysis for Multi-block Data: Common and Individual Feature Extraction}
{\author{Guoxu~Zhou, Andrzej~Cichocki ~\IEEEmembership{Fellow,~IEEE}, Yu Zhang, and Danilo~Mandic ~\IEEEmembership{Fellow,~IEEE}
\thanks{Manuscript received ........ This work was supported by the National Natural Science Foundation of China (grants 61103122, 61202155, 61333013, and U1201253),  the Guangdong Natural Science Foundation (2014A030308009),  and the JSPS KAKENHI (Grant No. 26730125).
}
\thanks{ Guoxu Zhou is with the Laboratory for Advanced Brain Signal Processing, RIKEN, Brain Science Institute, Wako-shi, Saitama 3510198, Japan. E-mail: zhouguoxu@brain.riken.jp.}
\thanks{ Andrzej~Cichocki is with the  RIKEN, Brain Science Institute, Wako-shi, Saitama 3510198, Japan,  and Systems Research Institute Polish Academy of Science, Warsaw, Poland. E-mail: cia@brain.riken.jp.}
\thanks{Yu Zhang is with the Key Laboratory for Advanced Control and Optimization for Chemical Processes, Ministry of Education, East China University of Science and Technology, Shanghai 200237, China (e-mail: yuzhang@ecust.edu.cn.}
\thanks{ Danilo~Mandic is with the Communication and Signal Processing Research Group, Department of Electrical and Electronic Engineering, Imperial College, London, United Kingdom. E-mail: d.mandic@imperial.ac.uk.}
}

}

\markboth{IEEE TRANSACTIONS ON NEURAL NETWORKS AND LEARNING SYSTEMS}
{ZHOU \MakeLowercase{\textit{et al.}}: Group Component Analysis for Multi-block Data: Common and Individual Feature Extraction}

\maketitle

\begin{abstract}
Real world data are often acquired as a collection of matrices rather than as a single matrix. Such multi-block data are naturally linked and typically share some common features and at the same time exhibit their own individual features, reflecting the background in which they are measured and collected. To exploit the linked nature of data, we propose a new framework for common and individual feature extraction (CIFE) which identifies and separates the common and individual features from multi-block data. Two efficient algorithms termed common orthogonal basis extraction (COBE) are proposed to extract the common basis which is shared by all data, independent on whether the number of common components is given or not. Feature extraction is then performed on the common and the individual subspaces separately, by incorporating dimensionality reduction and blind source separation techniques. Extensive experimental results on both synthetic and real-world data show significant advantages of the proposed CIFE method in comparison to the state-of-the-art.
\end{abstract}

\begin{IEEEkeywords}
Linked blind source separation, common and individual feature extraction, classification, clustering
\end{IEEEkeywords}

\IEEEpeerreviewmaketitle

\section{Introduction and Motivation}
\IEEEPARstart{T}{he} emergence of high-dimensional data structures requires new data analysis tools to be able to deal with the many aspects of this multifaceted problem, from data representation and interpretation to information retrieval. In this context, multi-block data analysis techniques are particularly interesting, as they accommodate multiple measurements of the same phenomenon under various experimentation conditions. For example, human electrophysiological signals in response to a certain stimulus, but from different subjects and trials, can be grouped together and naturally linked as multi-block data. Such data blocks share common information, and at the same time they also allow for individual data features to be kept.
Intuitively, this common shared information should help to discover connections between members of a data ensemble and can be used to {characterize this \emph{data ensemble}}, while the individual features may help to recognize or identify each \emph{individual member} of the data ensemble. The identification and separation of such common and individual information in order to employ the features highly relevant to the data analysis task at hand promises to significantly improve data analysis \cite{JIVE2013,BayesianJoint2014,DistComon2014,ComonCluster2013}. For example, shared features among tasks have been exploited to improve the performance of supervised and semi-supervised learning \cite{multiTask2007,Ando:2005}. In this paper, we  focus on an unsupervised  learning framework for the common and individual features across multi-block data. 

Common data analysis techniques that relate two data sets include canonical correlation analysis (CCA) which aims to maximize the correlation between two data sets \cite{CCA,lsCA2012tpami, onlineCCA} and the class of partial least squares (PLS) methods that maximize the cross-covariance  \cite{PLS1985, OnPLS2013, Trygg.Wold2003O2-PLS}.  CCA has also been generalized to multiple data sets in the context of blind source separation (BSS) and feature extraction \cite{mCCA,JBSSCCA,mLSCCA,mCCA:Witten,IJNS_COBE}.  For images, the population value decomposition (PVD) jointly analyzes same-size images \cite{PVD2011}, and can be considered as a special case of tensor (Tucker) decompositions, an active research topic in high-order data analysis and exploration  \cite{nmf-book}. Very recently, group independent component analysis (ICA) and independent vector analysis (IVA) were proposed to capture group variables from multi-block data \cite{AdaliIVASPM,gigICA2013,gIVA2013}. Common to these methods is that they  account for only correlated features within multi-block data, which limits their practical applications. The joint and individual variation explained (JIVE)  \cite{JIVE2013}  is a step in the right direction which simultaneously extracts both joint and individual variations across the members of a heterogeneous data ensemble. However, for the potential of JIVE to be fully exploited several issues need to be further addressed: (i) when common components are relatively weak, JIVE often gives principal components rather than true common components, thus compromising the fidelity of the extraction of true common components; (ii) rigorous quantitative analysis regarding the extracted common components is still missing; (iii) for heterogeneous data, JIVE has been mainly used for revealing gene-miRNA associations, but its potential in general machine learning applications remains unclear; iv) ways to improve data analysis performance by incorporating well-established component analysis tools also need further investigation.

To help resolve these issues, we propose a general framework for common and individual feature extraction (CIFE) for multi-block data, and provide:
\begin{enumerate}
  \item
  New efficient algorithms for common orthogonal basis extraction (COBE) from multi-block data. These new algorithms guarantee the identification of \emph{true common components} and can be applied to separate common and individual subspaces of multi-block data;

  \item 
{A unifying framework for multi-block data analysis, which deals with heterogeneous data structures by respectively applying suitable well-established matrix factorization methods (such as blind source separation (BSS) \cite{Cichocki2002} and nonnegative matrix factorization (NMF) \cite{nmf-book}) to the common and individual subspaces separately rather than to the global space of a data ensemble. This allows for more effective information retrieval and feature extraction for multi-block data.}

\item Two generic applications of CIFE---classification and clustering, illustrating the extent to which a simultaneous use of common and individual features can improve the performance in practical applications.
\end{enumerate}

 We also provide an in-depth analysis of the links between  COBE  and related methods such as CCA and  principal component analysis (PCA).
The analysis shows that the introduced common feature extraction can be interpreted as higher-order correlation analysis, where PCA is performed on the common subspace shared by all the available data rather than on a single global data set, as in ordinary PCA. 

The rest of the paper is organized as follows. In Section 2 the COBE method is discussed, including the problem statement, model, algorithms, and its links with CCA and other related methods. In Section 3 a general framework for CIFE is presented. The applications of CIFE in classification and clustering are presented in Section 4. Section 5 provides simulations on both synthetic and real-world data, verifying the proposed methods. Finally,  concluding remarks and directions for future work are provided in Section 6.

A conference summary of part of our results has been accepted for publication at ICASSP 2015 \cite{CIFA_ICASSP2015}. Relative to \cite{CIFA_ICASSP2015}, this journal version includes detailed derivations, new algorithms, potential applications in machine learning, and further experiments.

\section{Common Orthogonal Basis Extraction}
Consider a set of matrices $\mathcal{Y}=\{\mats{Y}\in\Real^{D\times J_n}: n\in\Set{N}\}$, $\Set{N}=\set{1,2,\ldots,N}$, and the following matrix factorization problem, whereby for each matrix $\mats{Y}\in\Set{Y}$, we seek
\begin{equation}
\label{eqMixing}
  \min_{\mats{A},\;\mats{B}} \frob{\mats{Y}-\mats{A}\mats{B}^T}^2, \quad n\in\Set{N},
\end{equation}
where the $R_n$ columns of $\mats{A}\in\Real^{D\times R_n}$ represent the latent variables in \mats{Y} (sources, bases, loading, etc), and $\mats{B}\in\Real^{J_n\times R_n}$ denotes the corresponding coefficient matrix (mixing, encoding, etc). We assume that $R_n < \min(D,J_n)$, which implies that $\mats{A}\mats{B}^T$ provides a compact/compressed or low-rank representation of \mats{Y}.

\begin{figure}[t]
\centering
\includegraphics[width=.85\linewidth]{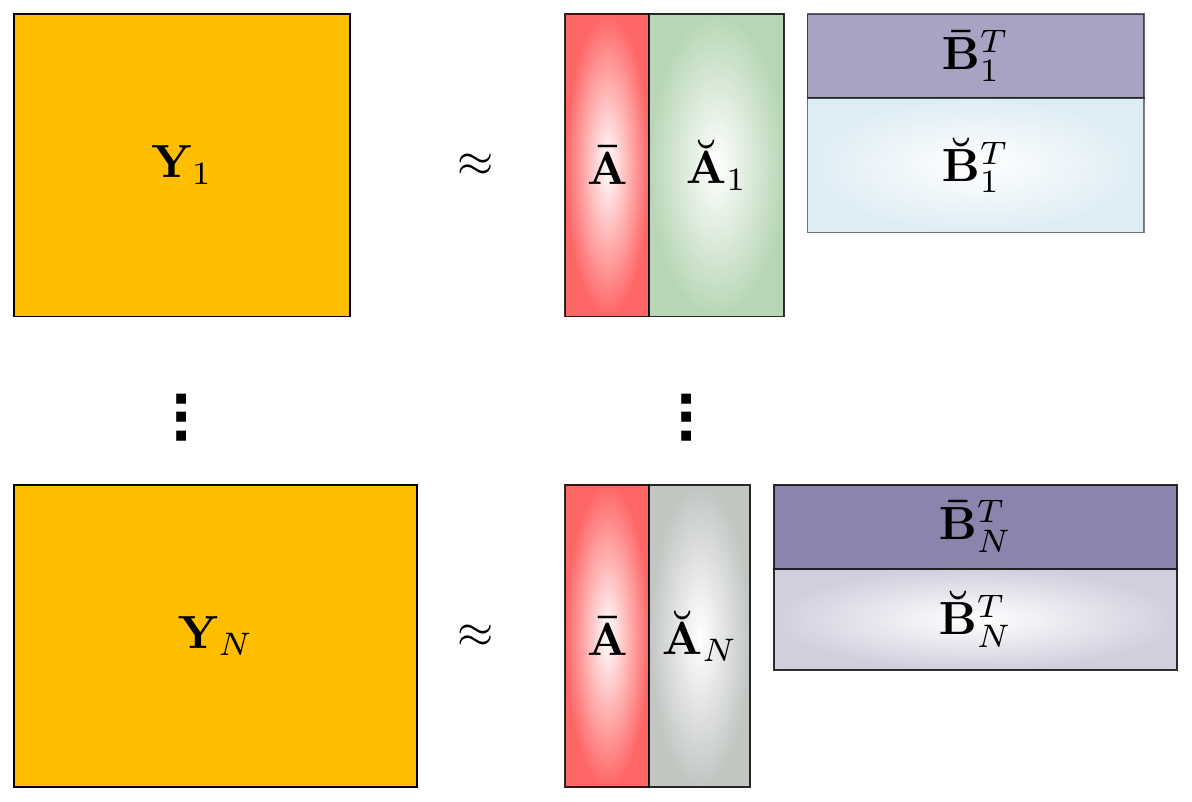}
\caption{The idea of common and individual feature extraction. The matrix \mat{\bar{A}} denotes the common (or highly correlated) features shared by all data while \mats{\breve{A}} are the individual features possessed by individual data blocks.}
\label{fig:figure1.pdf}
\end{figure}

A number of matrix factorization techniques exist to solve (\ref{eqMixing}), including PCA, BSS \cite{Cichocki2002}, however, these methods consider each matrix \mats{Y} separately. This is often both counter-intuitive and physically restrictive, since the members of the data ensemble \mats{Y} are likely to be naturally linked, thus sharing some common components. We therefore propose to perform a simultaneous analysis of the ensemble \Set{Y}, in order to obtain the latent variables in the form
\begin{equation}
\label{eqAstruct}
\mats{A}=\left[\mat{\bar{A}} \quad \mats{\breve{A}}\right], \quad n\in\Set{N},
\end{equation}
where  $\mat{\bar{A}}\in\Real^{D\times C}$, $\mats{\breve{A}}\in\Real^{D\times{(R_n-C)}}$, and $C\le\min\set{R_n:n\in\Set{N}}$. In other words, we assume that the sub-matrix $\mat{\bar{A}}$ contains the common components shared by all the matrices in \Set{Y} while the sub-matrix $\mats{\breve{A}}$ contains the individual information in each \mats{Y}. This allows us to factorize the data matrices in $\mathcal{Y}$ in a linked way, so that
\begin{equation}
  \label{eqLinkedMF}
  \begin{split}
    \mats{Y}\approx\mats{A}\mats{B}^T&=[\mat{\bar{A}} \quad \mats{\breve{A}}]\left[\begin{aligned} \mats{\bar{B}}^T \\ \mats{\breve{B}}^T \end{aligned} \right]  \\
    = & \mat{\bar{A}}\mats{\bar{B}}^T+\mats{\breve{A}}\mats{\breve{B}}^T \\
    \defeq& \mats{\bar{Y}}+\mats{\breve{Y}}, \quad n\in\Set{N},
  \end{split}
\end{equation}
where $\mats{\bar{B}}$ and $\mats{\breve{B}}$ are the partitions of the coefficients \mats{B} corresponding to $\mat{\bar{A}}$ and $\mats{\breve{A}}$. In this way, each data matrix \mats{Y} is represented through a combination of components from the common (shared) subspace $\mats{\bar{Y}}=\mat{\bar{A}}\mats{\bar{B}}^T$ and the individual (intrinsic) subspace $\mats{\breve{Y}}=\mats{\breve{A}}\mats{\breve{B}}^T$, as illustrated in \figurename \ref{fig:figure1.pdf}.

Our objective is to find the common and individual components $\mat{\bar{A}}$ and \mats{\breve{A}} which exhibit some desired properties from a given set of matrices \mats{Y}, $n\in\Set{N}$, without the knowledge of the mixing coefficients \mats{B} and possibly even without knowing the number of common components $C$. Recall that two special cases of (\ref{eqLinkedMF}) have been extensively studied in the past decades:
\begin{itemize}
  \item The case $C=0$, where no common components exist in \Set{Y} so that the problem boils down to factorizing each matrix $\mats{Y}\in\Set{Y}$ separately;
  \item The case $C=R_n$ for all $\mats{Y}$, which is equivalent to matrix factorization of a large global matrix created by stacking together all matrices \mats{Y}.
\end{itemize}
It is important to note that the solutions to \eqref{eqLinkedMF} are not unique. In fact, for any invertible matrix $\mat{\bar{Q}}$ of appropriate size, we have $\mats{\bar{Y}}=(\mat{\bar{A}\bar{Q}})(\inv{\bar{Q}}\mats{\bar{B}}^T)$ which is also a solution to (\ref{eqLinkedMF}). 

To reduce the solution space and to simplify the computation we shall consider the following three steps:

{\bf Step 1:} Consider the QR-decomposition of $\mat{\bar{A}}=\mat{UR}$ such that $\mat{{U}}^T\mat{{U}}=\matI$, where \matI{} is the identity matrix (we also use the symbol $\matI_C$ to denote the $C$-by-$C$ identity matrix). Upon substituting into (\ref{eqLinkedMF}), we obtain
\begin{equation}
\label{eqorthBase}
    \mats{A}\mats{B}^T= \begin{bmatrix}\mat{U} & \mats{\breve{A}}\end{bmatrix}
    \begin{bmatrix}\mat{R}\mats{\bar{B}}^T \\ \mats{\breve{B}}^T\end{bmatrix}, \quad n\in\Set{N},
\end{equation}
a comparison between (\ref{eqLinkedMF}) and (\ref{eqorthBase}) makes it possible to assume that  in \eqref{eqLinkedMF} $\mat{\bar{A}}^T\mat{\bar{A}}=\matI$, without loss of generality.

{\bf Step 2:} Since our aim is to separate ``shared" and ``individual" latent component subspaces, we can further assume that $\mat{\bar{A}}^T\mats{\breve{A}}=\matO, n\in\Set{N}$, where \matO{} is a zero matrix, which implies no interaction between the common and individual subspaces, \zgx{i.e. perfect separability of the common and individual features}. This assumption is reasonable and will not introduce any additional factorization error. To see this, consider
\begin{equation}
  \label{eqOrthA2}
  \mats{\breve{A}}\equiv\mat{\bar{A}\bar{A}}^T\mats{\breve{A}}+(\matI-\mat{\bar{A}\bar{A}}^T)\mats{\breve{A}}.
\end{equation}
Substituting  \eqref{eqOrthA2} into \eqref{eqLinkedMF}, we have
\begin{equation}
\label{eqCorthI}
\begin{split}
\mats{A}\mats{B}^T
              =& \mat{\bar{A}}\mats{\bar{B}}^T+\mats{\breve{A}}\mats{\breve{B}}^T \\
              =& \mat{\bar{A}}\mats{\bar{B}}^T+[\mat{\bar{A}\bar{A}}^T\mats{\breve{A}}+(\matI-\mat{\bar{A}\bar{A}}^T)\mats{\breve{A}}]\mats{\breve{B}}^T \\
              =& \mat{\bar{A}}[\mats{\bar{B}}^T+\mat{\bar{A}}^T\mats{\breve{A}}\mats{\breve{B}}^T]+[(\matI-\mat{\bar{A}\bar{A}}^T)\mats{\breve{A}}]\mats{\breve{B}}^T \\
\end{split}
\end{equation}
Upon comparing (\ref{eqCorthI}) and (\ref{eqLinkedMF}) and defining $\mats{\breve{A}}\defeq(\matI-\mat{\bar{A}\bar{A}}^T)\mats{\breve{A}}$ and $\mats{\bar{B}}\defeq\mats{\bar{B}}+\mats{\breve{B}}\mats{\breve{A}}^T\mat{\bar{A}}$, we arrive at $\mat{\bar{A}}^T\mats{\breve{A}}=\matO{}$, verifying that this assumption is not only reasonable, but also does not introduce any factorization error.

{\bf Step 3:} We consider the truncated singular value decomposition (SVD) of $\mats{\breve{A}}=\mats{U}\mats{\Lambda}\mats{V}^T$, where $\mats{U}^T\mats{U}=\matI{}$, $\mats{V}^T\mats{V}=\matI$, and $\mats{\Lambda}\in\Real^{(R_n-C)\times (R_n-C)}$ is invertible. Then $\mat{\bar{A}}^T\mats{\breve{A}}=\matO{}\Leftrightarrow\mat{\bar{A}}^T\mats{U}=\matO{}$. Setting $\mats{\breve{A}}\defeq\mats{U}$ and $\mats{\breve{B}}\defeq\mats{\Lambda}\mats{V}^T\mats{\breve{B}}$, we have $\mat{\bar{A}}^T\mats{\breve{A}}=\matO{}$ and $\mats{\breve{A}}^T\mats{\breve{A}}=\matI{}$.


Based on the above three steps for a known of number of common components $C$, the general factorization problem in \eqref{eqLinkedMF} can be reformulated as:
%
\begin{equation}
  \label{eqStdModel}
  \begin{split}
  \min_{\mat{\bar{A}},\mats{\breve{A}}} \quad & \sum_{n\in\Set{N}}\frob{\mats{Y}-\mat{\bar{A}}\mats{\bar{B}}^T-\mats{\breve{A}}\mats{\breve{B}}^T}^2, \\
    s.t.\quad
                & \mat{\bar{A}}^T\mat{\bar{A}}=\matI{}_C, \; \mats{\breve{A}}^T\mats{\breve{A}}=\matI_{R_n-C}, \\
                  \quad & \mat{\bar{A}}^T\mats{\breve{A}}=\matO, \quad n\in\Set{N}, \\
\end{split}
\end{equation}
where $\frob{\cdot}$ is the Frobenius norm of matrices. The JIVE method also solves the model \eqref{eqStdModel}, although this is not stated explicitly \cite{JIVE2013}. 


The factorization problem \eqref{eqStdModel} has a very close relationship with PCA, as it naturally boils down to standard low-rank approximation of matrices (ordinary PCA) when $R_n=C$, $\forall n\in\Set{N}$, so that  $\mat{\bar{A}}=\mat{A}$ can be found from
\begin{equation}
  \label{eqPPCA}
 \min_{\mat{A}}  \; \sum_{n\in\Set{N}}\frob{\mats{Y}-\mat{A}\mats{B}^T}^2, \quad
 s.t.  \;\; \mat{A}^T\mat{A}=\matI_{C}.
\end{equation}
In this case, if the data matrices \mats{Y} are stacked together to form a global $D\times \sum_{n}J_n$ matrix $\mat{\tilde{Y}}=\begin{bmatrix}\mats[1]{Y} & \mats[2]{Y} & \cdots \mats[N]{Y}\end{bmatrix}$, and similarly $\mat{\tilde{B}}=\begin{bmatrix}\mats[1]{B} & \mats[2]{B} & \cdots & \mats[N]{B}\end{bmatrix}$, then (\ref{eqPPCA}) can be viewed as a partitioned version of the global PCA of \mat{\widetilde{Y}}, that is
\begin{equation}
  \label{eqLPCA}
\min_{\mat{A}}  \; \frob{\mat{\tilde{Y}}-\mat{A}\mat{\tilde{B}}^T}^2, \quad
s.t.  \;\; \mat{A}^T\mat{A}=\matI_C.
\end{equation}
If \mat{\widetilde{Y}} is too large for a computer memory, we may resort to \eqref{eqPPCA} to perform PCA in practice.



However  when $C<R_n$, problem \eqref{eqStdModel} is not equivalent to PCA any more. The key difference between the models in \eqref{eqStdModel} and \eqref{eqPPCA} is  due to the  individual parts $\mats{\breve{A}}\mats{\breve{B}}^T$, that is, common components found by \eqref{eqStdModel} can also be interpreted as principal components of the common subspace $\mats{Y}-\mats{\breve{A}}\mats{\breve{B}}^T$. 
This clarifies that in each iteration the JIVE effectively performs joint PCA and individual PCA sequentially: (i) in the joint PCA step, the individual subspaces are removed prior to applying PCA to all data; (ii) the individual PCA step is performed on each single individual data after the common subspace has been removed. Since both the common and individual components are unknown, the JIVE method updates these two parts in an alternating manner.

\remark{Notice that in \eqref{eqStdModel} both \mats{A} and \mats{B} are need to be optimized, this is different from least squares where \mats{B} are fixed.}

\begin{figure}[t]
\centering
\includegraphics[width=.85\linewidth]{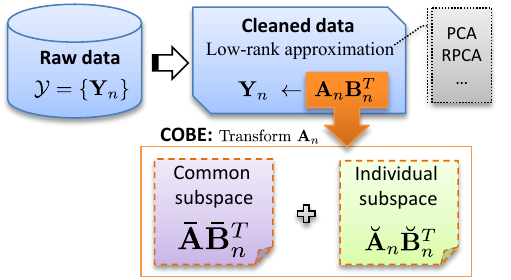}
\caption{Our two-step method for common and individual feature extraction. The separation of the common and individual subspaces via transforming \mats{A} will not introduce any additional factorization error.}
\label{fig:2stepCIFE}
\end{figure}

Since from \eqref{eqAstruct}-\eqref{eqCorthI} the separation of individual and common subspaces does not introduce any additional factorization error, by comparing \eqref{eqMixing} and \eqref{eqStdModel},   the matrix $\mats{A}=\begin{bmatrix}
\mat{\bar{A}} & \mats{\breve{A}}
\end{bmatrix}$ essentially gives the optimal rank-$R_n$ approximation of \mats{Y} with separated common and individual subspaces. This means that \eqref{eqStdModel} actually consists of two major functions: \emph{dimensionality reduction} (or low-rank approximation) of each single data matrix \mats{Y} and \emph{separation of common and individual components} of the data ensemble. The dimensionality reduction of \mats{Y} depends on each \mats{Y} only and its purpose is to  capture the variation as much as possible;  while the latter requires an integrated analysis of all data matrices with the purpose of extracting the common (or highly correlated) components shared by the data ensemble.  
The JIVE approach performs these two functions simultaneously  by applying the alternating least squares (ALS) to \eqref{eqStdModel} and quantifies the amount of  joint \emph{variation} between data types \cite{JIVE2013}.  In other words,  components extracted by JIVE are largely dominated by variances rather than correlations. As a result, the JIVE may fail to capture the common components with high correlations, especially when the common components in the data ensemble are relatively weak but are consistently present in all data sets.
In contrast, we think  it could be more natural to realize these two functions separately.
Hence, we consider a new two-step method to solve \eqref{eqStdModel}, which is also illustrated in \figurename \ref{fig:2stepCIFE}:

{\bf Step 1:} \emph{Dimensionality reduction:} update the matrices \mats{Y} in \eqref{eqStdModel} by their optimal rank-$R_n$ approximation $\mats{A}\mats{B}^T$ by solving \eqref{eqMixing} separately for each \mats{Y}. We call the original \mats{Y} \emph{raw} data and the reduced version $\mats{Y}\from\mats{A}\mats{B}^T$ \emph{cleaned} data;

{\bf Step 2:} \emph{Common and Individual Components Separation:} solve \eqref{eqStdModel} using the cleaned data.\\
Therefore, because the procedure  \eqref{eqAstruct}-\eqref{eqCorthI} that separates the common and individual subspaces  does not introduce any factorization error, in theory, we have
 \begin{equation}
\label{eqpcaCobe}
  \mats{Y}=\mats{A}\mats{B}^T=\mat{\bar{A}}\mats{\bar{B}}^T+\mats{\breve{A}}\mats{\breve{B}}^T, \;\forall n\in\Set{N},
\end{equation}
where \mats{Y} is the cleaned data.
Obviously, \eqref{eqpcaCobe} holds if and only if \mat{\bar{A}} contain only common components. This property can be further exploited to ensure the extraction of true common components, as detailed in Sections  \ref{subsec:COBE} and \ref{subsec:COBEc}.  Additional important advantages of this two-step method include: (i) the dimensionality reduction is more flexible in the sense that any suitable methods can be applied;  (ii) it can be performed in parallel when a large number of data sets are involved; and (iii)  it simplifies the subsequent separation of the common and individual components.

In the sequel, we shall assume that \mats{Y} in \eqref{eqStdModel} are the cleaned data processed by the above Step 1.


\subsection{The COBE Algorithm: The Number of Common Components $C$ is Unknown}
\label{subsec:COBE}
The estimation of the common components \mat{\bar{A}} plays a central role in solving \eqref{eqStdModel}. In fact, once \mat{\bar{A}} has been estimated, from \eqref{eqpcaCobe}, the coefficient matrices \mats{\bar{B}} can be computed from\footnote{If $\mats{Y}=\mat{\bar{A}}\mats{\bar{B}}^T+\mats{\breve{A}}\mats{\breve{B}}^T$ is exact as in \eqref{eqpcaCobe}, Equation \eqref{eqAc2Bc} is also exact. Otherwise \eqref{eqAc2Bc} is interpreted as the least squares solution of $\min \frob{\mats{Y}-\mat{\bar{A}}\mats{\bar{B}}^T-\mats{\breve{A}}\mats{\breve{B}}^T}^2$. Similar reasoning also applies to equations \eqref{eqLSA} and \eqref{eqA=QZ}. }
\begin{equation}
  \label{eqAc2Bc}
  \mats{\bar{B}}=\left(\mats{Y}^T-\mats{\breve{B}}\mats{\breve{A}}^T\right)\mat{\bar{A}}{(\mat{\bar{A}}^T\mat{\bar{A}})}^{-1}=\mats{Y}^T\mat{\bar{A}},\; n\in\Set{N},
\end{equation}
which then allows us to compute the common subspace $\mat{\bar{A}}\mats{\bar{B}}^T$ and the individual subspace $\mats{\breve{Y}}=\mats{Y}-\mat{\bar{A}}\mats{\bar{B}}^T$, $\forall n\in\Set{N}$, respectively.

To estimate \mat{\bar{A}} efficiently, from \eqref{eqpcaCobe} we have
\begin{equation}
  \label{eqLSA}
  \begin{bmatrix}\mat{\bar{A}} & \mats{\breve{A}}\end{bmatrix}=\mats{Y}\pinv{\mats{B}^T{}}, \quad \mats{A}^T\mats{A}=\matI_{R_n}, \quad n\in\Set{N},
\end{equation}
where $\mats{A}=\begin{bmatrix}\mat{\bar{A}} & \mats{\breve{A}}\end{bmatrix}$, $\mats{B}=\begin{bmatrix}\mats{\bar{B}} & \mats{\breve{B}}\end{bmatrix}$, and $\pinv{(\cdot)}$ denotes the Moore-Penrose matrix pseudo-inverse. In other words, by finding appropriate transformation matrices $\pinv{\mats{B}^T{}}$, we can obtain the desired common and individual component subspaces spanned by \mat{\bar{A}} and \mats{\breve{A}}, respectively. 

{\bf Computing the basis vectors of \mat{\bar{A}}}. Let $\mats{Y}=\mats{Q}\mats{H}$ such that $\mats{Q}^T\mats{Q}=\matI$; for each matrix \mats{Y} this only needs to be computed once by using, e.g., the QR decomposition or a truncated SVD of \mats{Y}. We next define $\mats{Z}\defeq\mats{H}\pinv{\mats{B}^T{}}$, so that \eqref{eqLSA} becomes
\begin{equation}
\label{eqA=QZ}
\begin{bmatrix}\mat{\bar{A}} & \mats{\breve{A}}\end{bmatrix}=(\mats{Q}\mats{H})\pinv{\mats{B}^T{}}=\mats{Q}\mats{Z}, \quad n\in\Set{N},
\end{equation}
and hence for any $n_1, n_2\in\Set{N}$, $n_1\neq n_2$, the following holds
\begin{equation}
\label{eqQz=a}
\left\{
   \begin{aligned}
  &\mats[n_1]{Q}\mats[n_1,k]{z}=\mats[n_2]{Q}\mats[n_2,k]{z}=\mat{\bar{a}}_k,      & \text{if }  k\le C;\\
  &\mats[n_1]{Q}\mats[n_1,k]{z}\neq \mats[n_2]{Q}\mats[n_2,k]{z}, & \text{if } k>C,
  \end{aligned}
  \right.
\end{equation}
where $\mats[n,k]{z}$ and \mats[k]{\bar{a}} are the $k$th columns of $\mats{Z}$ and \mat{\bar{A}}, respectively.
From (\ref{eqQz=a}), the first column of \mat{\bar{A}}, denoted by $\mat{\bar{a}}_1$, can be calculated by solving:
\begin{equation}
  \label{eqF1}
   \min_{\mat{\bar{a}}_1,\;\mats[n,1]{z}} f_1=  \sum_n\frob{\mats{Q}\mats[n,1]{z}-\mat{\bar{a}}_1}^2, \quad s.t. \;\;\mat{\bar{a}}_1^T\mat{\bar{a}}_1=1,
\end{equation}
where the objective function $f_1$ has to be very small in order to ensure that \mats[1]{\bar{a}} is a common basis vector, as governed by \eqref{eqpcaCobe} and \eqref{eqA=QZ}. Eq. \eqref{eqF1} can be minimized by using alternating least-squares (ALS), whereby by first fixing $\mats[n,1]{z}$ the optimal $\mat{\bar{a}}_1$ is found as
\begin{equation}
\label{eqc}
  \mat{\bar{a}}_1=\sum\nolimits_n{\mats{Q}\mats[n,1]{z}},
\end{equation}
which is then normalized to have a unit norm. Repeating for a fixed $\mat{\bar{a}}_1$ we obtain
\begin{equation}
\label{eqx}
  \mats[n,1]{z}=\mats{Q}^T\mat{\bar{a}}_1, \quad n\in\Set{N},
\end{equation}
and so on until convergence. A common column $\mat{\bar{a}}_1$ is considered to be found if $\min f_1\le\epsilon$ for a very small threshold $\epsilon\ge0$; otherwise, no common basis exists in \Set{Y}, and we terminate the iterations  \eqref{eqc} and \eqref{eqx}.

Upon finding a set of common basis vectors $\mats[1]{\bar{a}}, \mats[2]{\bar{a}}, \ldots,\mats[k]{\bar{a}}$, we need to ensure that repeated common basis vectors are not found when we seek the next common vector \mats[k+1]{\bar{a}}.  This is achieved by considering the following useful property of \mats{Z}. Let $\mats[n,C]{Z}\defeq\begin{bmatrix}
\mats[n,1]{z} & \mats[n,2]{z} & \ldots & \mats[n,C]{z}
\end{bmatrix}$, then from \eqref{eqA=QZ} we have
\begin{equation}
  \label{eqOrthZ}
  \mats[n,C]{Z}^T\mats[n,C]{Z}=\mats[n,C]{Z}^T\mats{Q}^T\mats{Q}\mats[n,C]{Z}=\mat{\bar{A}}^T\mat{\bar{A}}=\matI.
\end{equation}
In other words, $\mats[n,k+1]{z}^T\mats[n,k]{Z}=\matO$, and \mats[n,k+1]{z} is in the null space of $\mats[n,k]{Z}^T$, which allows us to update \mats{Q} as
\begin{equation}
\label{eqUpdateA}
\begin{split}
  \mats{Q}^{(k+1)}&=\mats{Q}(\matI-\mats[n,k]{Z}\mats[n,k]{Z}^T) \\
                            &=\mats{Q}^{(k)}(\matI-\mats[n,k]{z}\mats[n,k]{z}^T),
  \end{split}
\end{equation}
where $\mats{Q}^{(1)}=\mats{Q}$. This then yields $\mat{\bar{a}}_{k+1}$ through solving
\begin{equation}
  \label{eqFk}
  \begin{split}
     \min_{\mat{\bar{a}}_{(k+1)},\;\mats[n,k+1]{z}} f_{k+1}= & \sum_n\frob{\mats[n]{Q}^{k+1}\mats[n,k+1]{z}-\mat{\bar{a}}_{k+1}}^2, \\
    s.t. \quad &\mat{\bar{a}}_{k+1}^T\mat{\bar{a}}_{k+1}=1.
  \end{split}
\end{equation}
The minimum of $f_{k+1}$ can be obtained by repeating the procedure\footnote{For $k>1$ the matrix $\mats{Q}^{(k)}$ is not any more orthogonal. However, it can be verified that $\mats{Q}^{(k)}{}^T$ is the More-Penrose pseudo-inverse of $\mats{Q}^{(k)}$, thereby leading to the least squares solution $\mats[n,k]{z}=\mats{Q}^{(k)}{}^T\mats[1]{\bar{a}}$, i.e. \eqref{eqx}.} in solving \eqref{eqF1}. We distinguish between the following two cases:

\begin{enumerate}
  \item For $\min f_{k+1}\le\epsilon$, a new common basis vector $\mat{\bar{a}}_{k+1}$ is found. We then update $\mats{Q}^{(k+1)}$ using (\ref{eqUpdateA}) and then solve (\ref{eqFk}) to seek the next common basis vector.
  \item Otherwise, no common basis vector exists any more and a total of $C=k$ common orthogonal basis vectors are found as $\mat{\bar{A}}=\begin{bmatrix}\mat{\bar{a}}_1 & \mat{\bar{a}}_2 & \cdots & \mat{\bar{a}}_C\end{bmatrix}$.
\end{enumerate}
We refer to the above procedure for finding sequentially an orthogonal basis of the common space as the common orthogonal basis extraction (COBE), which is outlined in Algorithm \ref{algCobe}.

\begin{algorithm}
\caption{The COBE Algorithm}
\label{algCobe}
\begin{algorithmic}[1]
 \REQUIRE \mats{Y}, $n\in\Set{N}$, $\epsilon\ge0$.
 \STATE Let \mats{Y}=\mats{Q}\mats{H} such that $\mats{Q}^T\mats{Q}=\matI_{R_n}$ for all $n\in\Set{N}$.
\STATE $\mat{\bar{A}}=[\;]$, $\mats{Q}^{(1)}=\mats{Q}$, and $k=1$.
\WHILE {$f_k\le\epsilon$}
\WHILE {not converged}
\STATE $\mat{\bar{a}}_k=\sum_{n}\mats{Q}^{(k)}\mats[n,k]{z}/\frob{\sum_n\mats{Q}^{(k)}\mats[n,k]{z}}$;
\STATE $\mats[n,k]{z}=[\mats{Q}^{(k)}]^T\mats[k]{\bar{a}}$, $n\in\Set{N}$;
\ENDWHILE
\STATE $f_k=\sum_n\frob{\mats{Q}^{(k)}\mats[n,k]{z}-\mat{\bar{a}}_k}^2$;
\STATE $\mat{\bar{A}}=\begin{bmatrix}\mat{\bar{A}} & \mat{\bar{a}}_k\end{bmatrix}$;
\STATE $k=k+1$;
\STATE $\mats{Q}^{(k)}=\mats{Q}^{(k-1)}(\matI-\mats[n,k-1]{z}\mats[n,k-1]{z}^T)$, $n\in\Set{N}$.
\ENDWHILE
\RETURN $\mat{\bar{A}}=\begin{bmatrix}\mats[1]{\bar{a}} & \mats[2]{\bar{a}} & \cdots & \mats[C]{\bar{a}}\end{bmatrix}$, where $C=k-1$.
\end{algorithmic}
\end{algorithm}

\remark{The parameter $\epsilon$ controls the degree of similarity (i.e., correlation, in this paper) of the extracted components $\mats{Y}\pinv{\mats{Y}}\mats[k]{\bar{a}}\approx\mats[k]{\bar{a}}$. If $\epsilon=0$, the extracted components are exactly the same and equal to \mats[k]{\bar{a}}, otherwise, approximately identical  highly correlated components  are extracted (see Section 2.5 for a detailed discussion).}

\remark{The basis matrix \mat{\bar{A}} is not unique as $\mat{\bar{A}}\mat{U}$ forms another basis matrix that also spans the common subspace, here \mat{U} is an arbitrary orthogonal matrix with proper size.}

With the intuition that $f_k\approx 0$ if and only if $k\le C$ (see \eqref{eqQz=a}),  the threshold $\epsilon$ should be sufficiently small to  ensure that the extracted \mat{\bar{A}} does contain only common components. 
In practice, we may specify a very small value of $\epsilon$ such that the number of common components is overestimated. Then the Second ORder sTatistic of the Eigenvalues (SORTE)  method \cite{HeZ2010-2}, which was proposed to detect the gap between the eigenvalues corresponding to the signal space and those belonging to the noise space, can be applied here to estimate the number of common components. Suppose we have overestimated a total of $K$ `common' components \mats[k]{\bar{a}}, $k=1,2,\ldots,K$ with $R_n\ge K>C$, by using Algorithm \ref{algCobe}, and the corresponding errors are $f_k$ given in \eqref{eqFk} or re-evaluated as $f_k\defeq\frac{1}{N}\sum_{n=1}^N\frob{\mats{Y}\pinv{\mats{Y}}\mats[k]{\bar{a}}-\mats[k]{\bar{a}}}^2$. Then we apply  SORTE  to
\begin{equation}
f_1, f_2, \ldots, f_K, 
\end{equation}
to detect the gap between the common and individual subspaces, as illustrated in Section V and in the first simulation.

\subsection{The COBE Algorithm When The Number of Common Components $C$ is Given}
\label{subsec:COBEc}
For a given $C$, following the analysis in Section \ref{subsec:COBE}, the common components can be found by solving:
\begin{equation}
  \label{eqCobeII}
\min_{\mats{Z},\;\mat{\bar{A}}}\quad \sum_{n=1}^{N}\frob{\mats{Q}\mats{Z}-\mat{\bar{A}}}^2, \quad 
s.t. \;  \mat{\bar{A}}^T\mat{\bar{A}}=\matI,
\end{equation}
through alternating optimization with respect to \mats{Z} and \mat{\bar{A}}. In this way, when \mat{\bar{A}} is fixed, the optimal \mats{Z} is computed from
\begin{equation}
  \label{eqCobeIIx}
  \mats{Z}\from\mats{Q}^T\mat{\bar{A}},\; n\in\Set{N};
\end{equation}
while when \mats{Z}, $n\in\Set{N}$, are fixed, (\ref{eqCobeII}) is equivalent to
\begin{equation}
  \label{eqCobeIIc}
 \max_{\mat{\bar{A}}} \quad \text{trace}(\mat{P}^T\mat{\bar{A}}), \quad 
 s.t. \quad \mat{\bar{A}}^T\mat{\bar{A}}=\matI,
\end{equation}
where $\trace{\cdot}$ denotes the trace of a matrix and  $\mat{P}=\sum_{n=1}^{N}\mats{Q}\mats{Z}$.

To solve \eqref{eqCobeIIc}, let $\mat{P}=\mat{E\Lambda V}^T\in\Real^{D\times C}$ be the truncated SVD (tSVD) of \mat{P}, where $\mat{\Lambda}=\text{diag}(\lambda_1,\lambda_2,\ldots,\lambda_C)\in\Real^{C\times C}$ is a diagonal matrix with $\lambda_1\ge\lambda_2\ge\cdots\ge\lambda_C>0$. Motivated by the work\footnote{The main difference is that here $\mat{\bar{A}}$ is not necessarily square.} in \cite{MatrixComputations} (page 601),  the optimal solution of (\ref{eqCobeIIc}) becomes
\begin{equation}
  \label{eqCOBEIIa}
  \mat{\bar{A}}=\mat{EV}^T,
\end{equation}
which is easy to see by considering
\begin{equation}
\nonumber
\begin{split}
  \trace{\mat{P}^T\mat{\bar{A}}} & = \trace{\mat{V\Lambda E}^T\mat{\bar{A}}} =\trace{\mat{\Lambda}(\mat{E}^T\mat{\bar{A}V})}.
\end{split}
\end{equation}
In fact, as $\mat{E}^T\mat{E}=\matI$ and $(\mat{\bar{A}V})^T(\mat{\bar{A}V})=\matI$, we have $[\mat{E}^T\mat{\bar{A}V}]_{ii}\le1$, which means that $\trace{\mat{\Lambda}(\mat{E}^T\mat{\bar{A}V})}\le\sum_{i=1}^C\lambda_{i}$. Clearly, when $\mat{\bar{A}}=\mat{EV}^T$, then $\mat{\bar{A}}^T\mat{\bar{A}}=\matI$, $\mat{E}^T\mat{\bar{A}V}=\matI$ and $\trace{\mat{P}^T\mat{\bar{A}}}$ reaches its upper bound $\sum_{i=1}^C\lambda_{i}$.
We refer to this algorithm as the COBE$C$ and its pseudocode is given in Algorithm \ref{algCobeII}.

\begin{algorithm}[t]
\caption{The COBE$C$ Algorithm}
\label{algCobeII}
\begin{algorithmic}[1]
 \REQUIRE $C$ and \mats{Y}, $n\in\Set{N}$.
 \STATE Let \mats{Y}=\mats{Q}\mats{H} such that $\mats{Q}^T\mats{Q}=\matI_{R_n}$ for all $n$.
 \STATE Initialize \mats{Z} randomly.
\WHILE {not converged}
\STATE{$\mat{P}=\sum_{n\in\Set{N}}\mats{Q}\mats{Z}$. }
\STATE{\mat{\bar{A}}=$\mat{EV}^T$, where $[\mat{E},\;\mat{\Lambda},\;\mat{V}]=\text{tSVD}(\mat{P},C)$}.
\STATE{$\mats{Z}\from\mats{Q}^T\mat{\bar{A}}$}.
\ENDWHILE
\RETURN $\mat{\bar{A}}$.
\end{algorithmic}
\end{algorithm}

\subsection{Pre-processing: Dimensionality Reduction}
Like CCA, COBE is meaningless if $R_n=D$ for all $n$, since for any $D\times D$ invertible matrix $\mat{\bar{A}}$ there always exist matrices \mats{\bar{B}} such that $\mats{Y}=\mat{\bar{A}}\mats{\bar{B}}^T$, i.e., any $D\times D$ invertible matrix forms a common basis. For this reason in model \eqref{eqStdModel} the condition $R_n<D$ is required for all \mats{Y}. This requirement is not too restrictive and has physical justification, as for real-world data of high dimensionality, the latent rank is often significantly lower than the dimensionality of the observed data. If $R_n<D$ is not satisfied, we need to perform dimensionality reduction (such as PCA) by, e.g. solving \eqref{eqMixing} prior to applying COBE, which justifies the necessity of the proposed two-step method. However, this is not the only reason to apply dimensionality reduction---it also helps to reduce the computational complexity, and particularly, to reduce noise, outliers, and artifacts in the original raw data. After dimensionality reduction, we obtain cleaned data $\mats{Y}\approx\mats{A}\mats{B}^T$ with $\rank{\mats{A}}=R_n<D$, and can apply COBE to the cleaned data to obtain $\mats{A}\mats{B}^T=\mat{\bar{A}}\mats{\bar{B}}^T+\mats{\breve{A}}\mats{\breve{B}}^T$. Notice that if $\mats{A}\mats{B}^T$ is interpreted as the PCA of matrix \mats{Y}, COBE simply rotates/transforms the columns of \mats{A} so that the common subspace and the individual subspace are completely disjoint, as illustrated in \figurename \ref{fig:2stepCIFE}.



\remark{Standard PCA performs dimensionality reduction for i.i.d. Gaussian noise; for sparse distributions we may use robust PCA (RPCA) \cite{rPCA2011}, while the number of latent components $R_n$ can be estimated using, e.g. SORTE \cite{HeZ2010-2}, before applying PCA.}

\subsection{Relation to Other Methods}
\begin{figure}[!t]
\centering
\includegraphics[width=.5\textwidth]{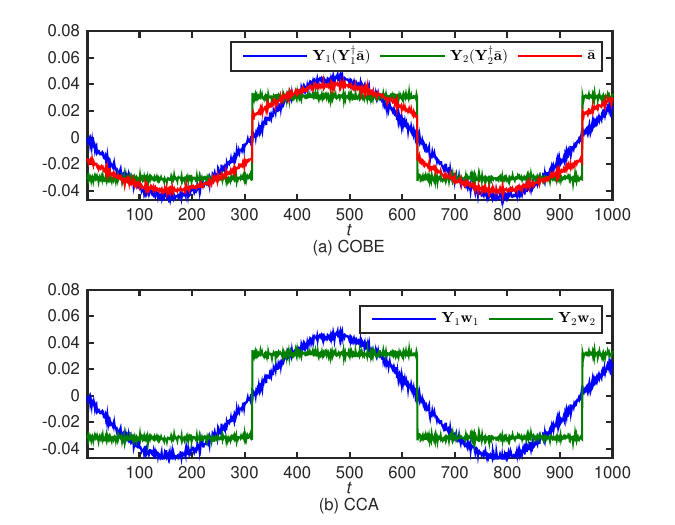}
\caption{(a) Illustration of how  COBE was able to capture the common information between two data sets. Here \mat{\bar{a}} can be interpreted as the principal component of the normalized common (or highly correlated)  components $\mats[n]{Y}({\mat{Y}}_n^{\dagger}\mats{\bar{a}})$, $n=1,2$. (b) The canonical variables (i.e., $\mats{Y}\mats[n]{w}$, $n=1,2$) obtained by using  CCA.}
\label{figcobecca}
\end{figure}

To illustrate the relation between COBE and CCA, recall that for data matrices  $\mats[1]{Y}$ and $\mats[2]{Y}$, CCA finds the vectors \mats[1]{w} and \mats[2]{w} which maximize the correlation $\rho=\text{corr}(\mats[1]{Y}\mats[1]{w},\mats[2]{Y}\mats[2]{w})$, while COBE extracts only components for which the correlation is higher than a specified threshold.  We next shed further light on this relationship by introducing the following bound on the correlations of the variables involved.
\begin{Proposition}
\label{th::cca}
   Let \mats{Y} be row-centered (i.e., with zero mean) random variables. Suppose that $\frob{\mats{Y}\mats{w}-\mat{\bar{a}}}\le\epsilon<\sqrt{2}-1$ with $\frob{\mat{\bar{a}}}=1$, $\forall n\in\Set{N}$, then for $\forall m,n\in\Set{N}$, we have
  \begin{equation}
    \text{corr}(\mats[m]{Y}\mats[m]{w},\mats{Y}\mats{w})\ge \frac{1-(2\epsilon+\epsilon^2)}{1+(2\epsilon+\epsilon^2)}>0.
  \end{equation}
\end{Proposition}
{\begin{IEEEproof}
  \label{apdCCA}
From $\frob{\mat{\bar{a}}}=1$ and $\frob{\mats{Y}\mats{w}-\mat{\bar{a}}}\le\epsilon<1$, we have
\begin{equation}
\label{eqCorrNorm}
  0<1-\epsilon\le\frob{\mats{Y}\mats{w}}\le 1+\epsilon, \; \forall n\in\Set{N}.
\end{equation}
Moreover, $\forall m, n\in\Set{N}$, the following holds
\begin{equation}
\begin{aligned}
      &\frob{\mats[m]{Y}\mats[m]{w}-\mats{Y}\mats{w}} \\
  \le & \frob{\mats[m]{Y}\mats[m]{w}-\mat{\bar{a}}}+\frob{\mats{Y}\mats{w}-\mat{\bar{a}}} \\
  \le & 2\epsilon.
\end{aligned}
\end{equation}
Hence,
\begin{equation}
\label{eqCorr1}
\begin{aligned}
    & 2\mats[m]{w}^T\mats[m]{Y}^T\mats{Y}\mats{w} \\
  =&\frob{\mats[m]{Y}\mats[m]{w}}^2+\frob{\mats{Y}\mats{w}}^2-\frob{\mats[m]{Y}\mats[m]{w}-\mats{Y}\mats{w}}^2 \\
  \ge &2(1-\epsilon)^2-4\epsilon^2,
\end{aligned}
\end{equation}
and from (\ref{eqCorrNorm}) and (\ref{eqCorr1}), we have

\begin{equation}
\notag
\begin{split}
  \text{corr}(\mats[m]{Y}\mats[m]{w},\mats{Y}\mats{w})& =\frac{\mats[m]{w}^T\mats[m]{Y}^T\mats{Y}\mats{w}}{\frob{\mats[m]{Y}\mats[m]{w}}\frob{\mats{Y}\mats{w}}}\\
  & \ge \frac{1-2\epsilon-\epsilon^2}{\frob{\mats[m]{Y}\mats[m]{w}}\frob{\mats{Y}\mats{w}}} \\
  & \ge \frac{1-2\epsilon-\epsilon^2}{(1+\epsilon)^2} \\
  & = \frac{1-2\epsilon-\epsilon^2}{1+2\epsilon+\epsilon^2}. \\
  \end{split}
\end{equation}
\end{IEEEproof}}
In other words, if $f_i$ in (\ref{eqF1}) and (\ref{eqFk}) are upper bounded, this, in turn, leads to lower bounded correlations between the projected variables $\mats{Y}\mats{w}$, $n\in\Set{N}$. Since $\text{corr}(\mats[1]{Y}\mats[1]{w},\mats[2]{Y}\mats[2]{w})\to 1$ as $\epsilon\to0$,  COBE can be interpreted as a higher order correlation analysis (HCA) method. \figurename \ref{figcobecca} illustrates the difference in operation between COBE and CCA for multiple data sets. In this simulation we first generated two matrices $\mats{S}\in\Real^{1000\times10},\; n=1,2$. The first column of \mats[1]{S} was $\mats[1,1]{s}(t)=\text{sin}(0.01t)$, and that of \mats[2]{S} was $\mats[2,1]{s}(t)=\text{sign}(\mats[1,1]{s}(t))$, $t=1,2,\ldots,1000$. Other entries were drawn from independent standard normal distributions. The matrices \mats[n]{S} were mixed via different coefficient matrices $\mats[n]{M}\in\Real^{10\times10}$, $n=1,2$, whose entries were drawn from independent standard normal distributions such that $\mats{Y}=\mats{S}\mats{M}^T$ ($n=1,2$). Apparently, $\mats[1,1]{s}(t)$ and $\mats[2,1]{s}(t)$ are highly correlated. Hence, by applying CCA we obtained the first pair of canonical variables $\mats[n,1]{\hat{s}}(t)=\mats[n]{Y}\mats[n]{w}$ with the correlation $\text{corr}(\mats[1,1]{\hat{s}}(t),\mats[2,1]{\hat{s}}(t))=0.8867$, where \mats[n]{w} were the corresponding canonical coefficients, $n=1,2$. The red line in \figurename \ref{figcobecca}(a) shows the common components \mat{\bar{a}} extracted by COBE, together with the blue and green lines corresponding highly correlated components extracted from \mats{Y}, i.e., $\mats[n]{Y}({\mat{Y}}_n^{\dagger}\mat{\bar{a}})$, $n=1,2$, which matched very well the canonical variables obtained by CCA (For this reason, we also call $\mats[n]{Y}({\mat{Y}}_n^{\dagger}\mat{\bar{a}})$ normalized common components). However, COBE will  extract only the components with very high correlations, as stated in Proposition \ref{th::cca}. From the figure, $\mat{\bar{a}}$ can be interpreted as the principal component of the normalized common components, the information that is not provided in CCA. 

\begin{figure}[!th]
    \centerline{
    \subfloat[PCA]{
    \includegraphics[height=1.2in]{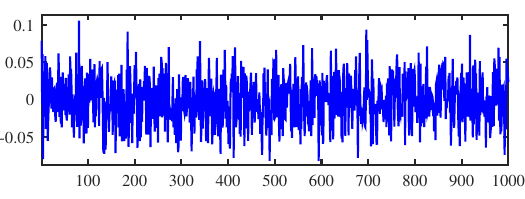}}}
    \centerline{
    \subfloat[JIVE]{
    \includegraphics[height=1.2in]{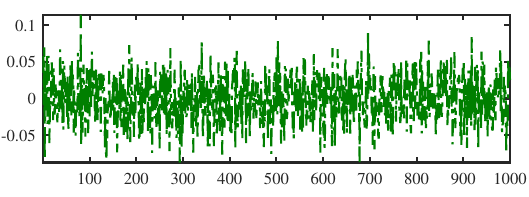}}}
    \centerline{
    \subfloat[COBE]{
    \includegraphics[height=1.2in]{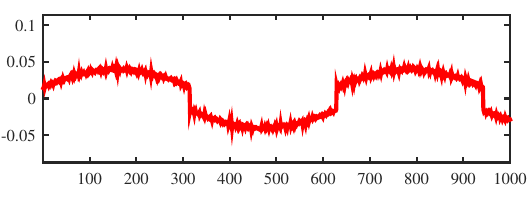}}}
\caption{Relation between  COBE,  JIVE, and  PCA:   COBE  extracted the  principal component of  normalized common (or highly correlated) columns whereas  PCA gave principal components of all columns, and  JIVE captured the joint variation and hence failed to extract the highly correlated components with relatively weak energy.}
\label{figcobepca}
\end{figure}

Further to showing in \eqref{eqStdModel} and \eqref{eqPPCA} that COBE has a close relation with PCA, \figurename \ref{figcobepca} illustrates the difference between COBE, JIVE and PCA using the same data as in \figurename \ref{figcobecca}. The principal component was computed from a concatenated version of \mat{\tilde{Y}}, defined in \eqref{eqLPCA}. Basically, COBE identifies the principal components \mat{\bar{A}} of normalized  common components (corresponding to the canonical variables in CCA), whereas PCA seeks the principal components of  the global data set \Set{Y}, and  JIVE captures the \emph{joint variation}. In this example, JIVE gave principal components of sorts, rather than common components, illustrating that the power of components dominates the results obtained by JIVE, see \figurename \ref{figcobepca}. In this sense, COBE is closer to CCA while JIVE is closer to PCA (and PLS); COBE can also be viewed as a regularized version of JIVE. In terms of computational demands, compared with JIVE which in the computation of the common subspace involves frequent SVDs of  huge matrices containing all data, COBE (COBE$C$) is more efficient in the optimization and more physically intuitive and flexible in the estimation of number of common components.


\subsection{Scalability For Large-Scale Problems}
For large-scale data the indices $D$ and $J_n$ ($n\in\Set{N}$) in \eqref{eqMixing} are quite large. First, recall that in \eqref{eqF1}, \eqref{eqFk}, and \eqref{eqCobeII} we use the dimensionality reduced matrices $\mat{Q}\in\Real^{D\times R_n}$ with $R_n<D$, and hence the value of $J_n$ is generally not an issue. On the other hand, for a very large $D$,  the time and memory requirements of COBE can be reduced in the following way. Let $\mat{P}\in\Real^{D_{\text{P}}\times D}$ be a random matrix with $\max_{n\in\Set{N}}(R_n)<D_{\text{P}}\ll D$. Then, from \eqref{eqLSA} we can first solve the model:
\begin{equation}
  \label{eqPCobe}
  \min \sum_{n\in\Set{N}}\frob{\mats{Y}^{\text{P}}\mats{W}-\mat{\bar{A}}^{\text{P}}}^2
\end{equation}
where $\mats{Y}^{\text{P}}=\mat{P}\mats{Y}\in\Real^{D_{\text{P}}\times J_n}$ is much smaller than \mats{Y}, and $\mat{\bar{A}}^{\text{P}}=\mat{P}\mat{\bar{A}}$. After the matrices \mats{W} have been estimated by using COBE or COBE$C$, the corresponding common basis can be computed from $\mat{\bar{A}}=\mats{Y}\mats{W}$. Obviously, $\mat{P}\mats{Y}\mats{W}=\mat{P}\mat{\bar{A}}$ as long as $\mats{Y}\mats{W}=\mat{\bar{A}}$, in other words, no common basis vectors are lost. In the worst case, this approach may give fake common components $\mat{\bar{a}}_k$ when  $\mats{Y}\mats[n,k]{w}-\mat{\bar{a}}_k$ occasionally lies in the null space of $\mat{P}$. In practice, this is not an issue, as these fake common components can be easily detected by examining the value of $\frob{\mats{Y}\mats[n,k]{w}-\mat{\bar{a}}_k}^2$.

\begin{figure*}[!t]
\centering
  \includegraphics[width=.9\textwidth]{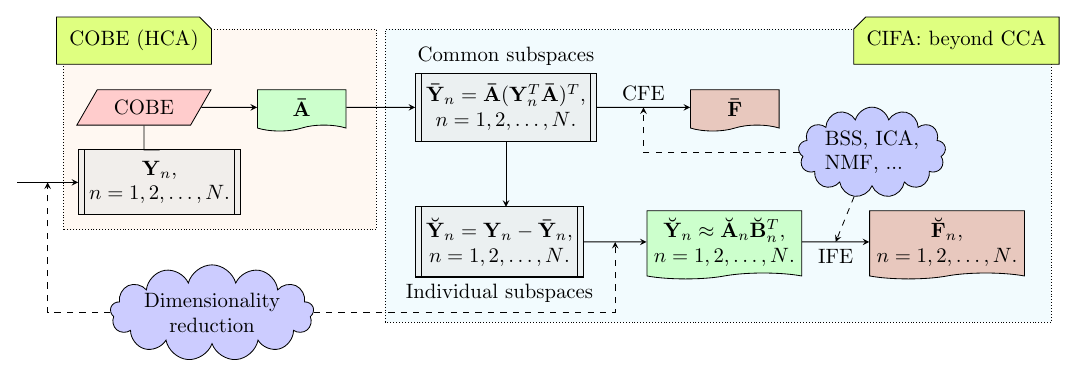}\\
  \caption{General framework of common and individual feature extraction (CIFE), i.e., group  component analysis from multi-block data.}
  \label{figFlowDia}
\end{figure*}

\section{Common And Individual Feature Extraction (CIFE)}
We shall now illuminate the versatility of the COBE approach over several established feature extraction paradigms.
\subsection{Linked BSS with Pre-whitening}
\label{sec:LinkedBSS}
We have so far considered the orthogonal components \mat{\bar{A}}, this however does not guarantee unique common components as the columns of $\mat{\bar{A}U}$ also form a common orthogonal basis for any orthogonal matrix \mat{U}. If our aim is to project the common components onto a feature space with some desired properties (uniqueness), this can be achieved by BSS \cite{Cichocki2002}, which finds latent variables \mat{S} from their linear mixtures $\mat{Y}=\mat{SM}^T$ such that
\begin{equation}
  \label{eqBSS}
  \mat{\hat{S}}=\Psi(\mat{Y})=\mat{SP\Lambda}
\end{equation}
where $\Psi$ denotes a suitable BSS algorithm, \mat{M} is the unknown mixing matrix, while \mat{P} and \mat{\Lambda} are a permutation matrix and a scaling matrix, respectively, that model the unavoidable ambiguities of BSS. Assuming that the latent sources \mat{\bar{F}} satisfy
\begin{equation}
\label{eqLinkedBSS}
  \mats{\bar{Y}}=\mat{\bar{F}}\mats{\bar{M}}^T,
\end{equation}
where $\mats{\bar{Y}}=\mat{\bar{A}}\mats{\bar{B}}^T$, we have
\begin{equation}
\label{eqLBSSWhitening}
  \mat{\bar{A}}=\mat{\bar{F}}(\pinv{\mats{\bar{B}}}\mats{\bar{M}})^T,
\end{equation}
 illustrating that the columns of \mat{\bar{A}} are simply  linear mixtures of the sources \mat{\bar{F}}, so that \mat{\bar{F}} can be estimated via BSS as
\begin{equation}
  \mat{\bar{F}}=\Psi(\mat{\bar{A}}).
\end{equation}
In this case \mat{\bar{A}} is a pre-whitened version of \eqref{eqLinkedBSS}, which stems from \eqref{eqLBSSWhitening} and the fact that $\mat{\bar{A}}^T\mat{\bar{A}}=\matI$. The major advantage of using BSS in this context is that we may obtain common features which exhibit some  desired properties such as sparsity, independence, nonnegativity. This is achieved by imposing appropriate constraints and penalties on \mat{\bar{F}}. Moreover, we may extract even more than $C$ common signals from \mat{\bar{A}} or \mats{\bar{X}}, which is a challenging problem referred to as underdetermined BSS. To this end, for example, we can apply the  novel tensor based approach proposed in \cite{TNN-TFUBSS}. This method advanced the study of this topic in the sense that  it can exactly recover  as many as $2C-1$ sources from $C$ observations at every autoterm time-frequency points, no matter how many active sources there are, which is so far the state-of-the-art for the separation of nonstationary sources.  In summary, the BSS procedure adds significantly increased versatility and flexibility to the COBE approach. We refer to the above procedure that combines COBE and BSS  as linked BSS to indicate that we perform BSS  on multi-block linked data \mats{Y}.

Note that the JBSS method in \cite{JBSSCCA} also performs BSS involving multi-block data; it extracts a group of signals with the highest correlations each time, and requires that the extracted groups have distinct correlations. In other words, the JBSS method can be viewed as a way to realize BSS by applying multiple-set CCA. In contrast, the proposed linked BSS method extracts a common basis first, it then applies ordinary BSS to discover common components with some desired properties and diversities. 
Recently in neural science, group ICA and independent vector analysis (IVA) have been widely applied to capture common (group) variables or components from a group (set) of data matrices \cite{groupICA2008guo,gigICA2013,gIVA2013}. These approaches are somewhat related to our proposed linked BSS model, however, they apply ICA to a global data space, which implicitly assumes that all data matrices are spanned by only common statistically independent components. The linked BSS differs from these methods because: (i) taking into account that the data matrices in a group not only share some common components but may also  contain individual components, the linked BSS can capture more reliable group variables as it performs BSS on the common subspace instead of on a global data space; (ii) the linked BSS is capable of capturing  components with various diversities or properties. In other words, within the linked BSS we may apply not only ICA, but also NMF (see Section 3.2) or any other suitable component analysis methods. These two distinguishing properties make our linked BSS more flexible and versatile and allow us to extract more physically meaningful components.


\subsection{Common Nonnegative Features Extraction (CNFE)}
For nonnegative latent sources  $\mat{\bar{F}}$, we cannot apply NMF methods on \mat{\bar{A}} directly. Instead, we need to first extract the common subspace using \eqref{eqAc2Bc}, i.e., $\mats{\bar{Y}}=\mat{\bar{A}\bar{A}}^T\mats{Y}$, and subsequently use the following low-rank approximation based (semi-) nonnegative matrix factorization (NMF) model \cite{lraNMF2012}:
\begin{equation}
  \label{eqCNCP}
 \min \; \sum_n\frob{\mat{\bar{F}}\mats{\bar{M}}^T-\mat{\bar{A}}\mats{\bar{B}}^T}^2, \; s.t. \quad  \mat{\bar{F}}\matge\matO.
 \end{equation}
The subsequent use of low-rank NMF (if $\mats{\bar{M}}$ is also nonnegative)  or low-rank semiNMF (where $\mats{\bar{M}}$ is real-valued) allows us to extract the common nonnegative components $\mat{\bar{F}}$. For example, the following iterative multiplicative update rules produces nonnegative components \mat{\bar{F}} and \mat{\bar{M}}:
\begin{equation}
  \label{eqCNFE}
  \begin{aligned}
    \mat{\bar{F}} &\from \mat{\bar{F}}\hdp \matdiv{[\mat{\bar{A}}\sum_n(\mats{\bar{B}}^T\mats{\bar{M}})]_+}{\mat{\bar{F}}(\sum_n\mats{\bar{M}}^T\mats{\bar{M}})}, \\
   \mats{\bar{M}} &\from \mats{\bar{M}}\hdp \matdiv{[\mats{\bar{B}}(\mat{\bar{A}}^T\mat{\bar{F}})]_+}{\mats{\bar{M}}(\mat{\bar{F}}^T\mat{\bar{F}})}, \; n\in\Set{N},
  \end{aligned}
\end{equation}
where $\hdp$ and $\matdiv{\mat{}}{\mat{}}$ are element-wise product and division of matrices, see \cite{lraNMF2012} for detailed convergence analysis. Similar to BSS, we may impose additional constraints to extract unique nonnegative components, see \cite{TNN-MVCNMF,YangNMFdetTNNLS, YangSNBSS2012, XieNMFTCSI}.

\subsection{Individual Feature Extraction (IFE)}
Besides the above common features \mat{\bar{F}} or \mat{\bar{A}} extracted using common feature extraction (CFE) methods, each data matrix also has its own \emph{individual} features contained in the matrix $\mats{\breve{Y}}=\mats{Y}-\mats{\bar{Y}}$; these are often helpful in classification and recognition. Notice that although \mats{\breve{Y}} is of the same size as \mats{Y}, it is rank deficient since $\rank{\mats{\breve{Y}}}+\rank{\mat{\bar{Y}}}=R_n$. Hence, dimensionality reduction of \mats{\breve{Y}} should be carefully addressed prior to further analysis. To estimate \mats{\breve{A}} and \mats{\breve{B}}, we can apply any standard dimensionality reduction method discussed in Section 2.4 (and the associated rank estimation techniques) to each $\mats{\breve{Y}}$ separately, followed by BSS or related methods to extract the features in $\mats{\breve{A}}$ and $\mats{\breve{B}}$. However, there is a major difference between the dimensionality reduction methods considered here and those in the pre-processing stage. In the pre-processing stage dimensionality reduction is rather general-purpose and relatively simpler, whereas here the dimensionality reduction is related to a specific task at hand. For example, we may wish to visualize  data in a low-dimensional space \cite{Bunte2011}, or to extract discriminative information, or to establish neighbor relationship. The above procedure is referred to as individual feature extraction (IFE), as the extracted features are only presented in individual datasets.

\figurename \ref{figFlowDia} shows the concept of the proposed common and individual feature extraction (CIFE).


\section{Example Applications}
\subsection{Classification Using Common Features}
In classification and pattern recognition, training data contain training samples and their labels, while the objects belonging to the same category naturally share some common features. More specifically, for a set of common features extracted from the $k$th category, $k\in\Set{K}=\{1,2,\ldots,K\}$, denoted by \mats[k]{\bar{F}}, upon arrival of a new test sample $\mats[t]{y}\in\Real^D$, we can compute its matching score $r_t(k)$ with each \mats[k]{\bar{F}} as:
\begin{equation}
  \label{eqMatchingScore}
  r_t(k)=\text{Matching}(\mats[t]{y},\mats[k]{\bar{F}}), \quad k\in\Set{K}.
\end{equation}
Since the samples within a certain class should share some features, the label of \mats[t]{y} is estimated as
\begin{equation}
  \label{eqClassifyLabel}
  l_t=\arg\max\limits_{k\in\Set{K}}r_t(k).
\end{equation}

\begin{figure}[!t]
\centering
  \includegraphics[width=.45\textwidth]{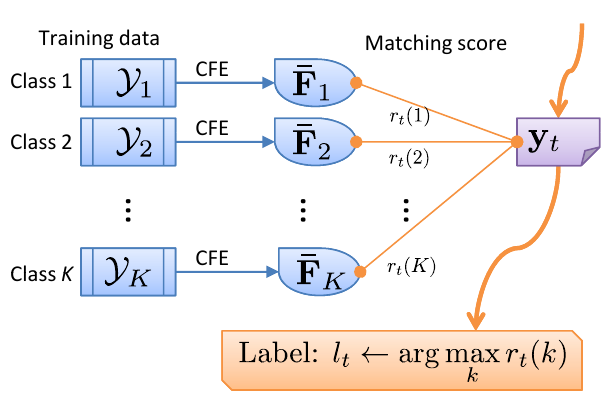}\\
  \caption{Classification  using common features extracted from each class of training data.}
  \label{figCFEClassify}
\end{figure}

The matching score $r_t(k)$ can be defined in many ways, such as the Euclidean distance or correlation (angle) between \mats[t]{y} and the space spanned by \mats[k]{\bar{F}}, which can be solved via least-squares and CCA, respectively. See \figurename \ref{figCFEClassify} for the diagram.

\subsection{Clustering Using Individual Features}
Cluster analysis assigns a set of objects to clusters in such a way that the objects belonging to the same cluster are most similar. Unlike classification, clustering employs an unsupervised learning approach. In the clustering analysis, it is usual that all the samples have some common features although they may be from different clusters. For example, in human face image analysis, every face has common facial features such as cheek, nose, eyes, and mouth, whose shapes and locations are similar. Their common features are not meaningful for clustering as they do not provide any discriminative information. It is therefore logical to first remove these common/similar features across all the samples and then to use their individual features to cluster the objects.

\begin{figure}[!t]
    \centerline{
    \subfloat[\mat{\bar{F}}]{
    \includegraphics[width=0.05\textwidth]{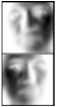}} \quad
    \subfloat[Examples of individual features $\mat{\breve{f}}_t$]{
    \includegraphics[width=0.32\textwidth]{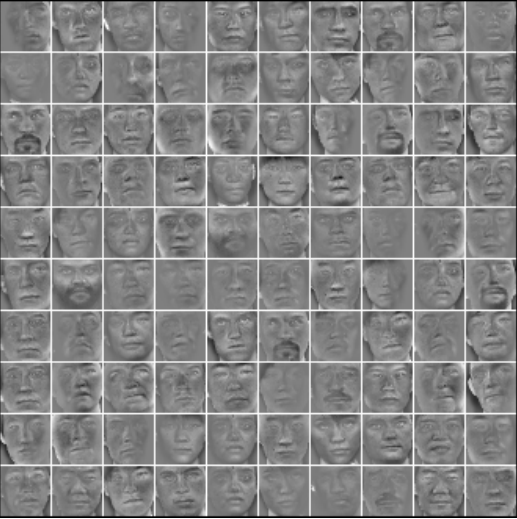}}
    }
    \caption{Extraction of the common features from the PIE database using COBE with CNFE. (a) \emph{Common faces}. (b) The first 64 samples of \emph{individual faces} obtained by removing the common components, observe a variety of local individual features.}
    \label{figPIEcifa}
  \end{figure}

\figurename \ref{figPIEcifa} shows that COBE incorporating CNFE is capable of extracting common faces (features) in the PIE database (details are given in the next section). We empirically set $C=2$ and used CNFE to extract common nonnegative components;  the common faces in \figurename \ref{figPIEcifa}(a) contain some basic features of human faces. On the other hand, \figurename \ref{figPIEcifa}(b) shows the accentuated \emph{individual local features}. These individual features are quite helpful to improve the accuracy of clustering and recognition. In our individual feature based clustering method we used the following steps:
\begin{enumerate}
  \item Randomly split the samples \mats[t]{y} into $N$ groups to construct \mats{Y}, where  $t\in\Set{T}=\{1,2,\ldots,T\}$ and $n\in\Set{N}$;
  \item Apply COBE to extract the common features \mat{\bar{A}} of $\{\mats{Y},\;n\in\Set{N}\}$;
  \item Remove the common features from \mats{Y} by letting $\mats{\breve{Y}}=\mats{Y}-\mats{\bar{Y}}=\mats{Y}-\mat{\bar{A}}\mat{\bar{A}}^T\mats{Y}$;
  \item Perform dimensionality reduction and feature extraction on $\begin{bmatrix}
  \mats[1]{\breve{Y}} & \mats[2]{\breve{Y}} & \cdots & \mats[N]{\breve{Y}}
  \end{bmatrix}$ to obtain the features $\mat{\breve{F}}=\begin{bmatrix}
   \mats[1]{\breve{f}} & \mats[2]{\breve{f}} & \cdots & \mats[T]{\breve{f}}
  \end{bmatrix}$;
  \item Apply clustering algorithms on $\{\mats[t]{\breve{f}}, \; t\in\Set{T}\}$, where \mats[t]{\breve{f}} corresponds to the original objects \mats[t]{y}.
\end{enumerate}

\section{Simulations and Validation}
{\bf Linked BSS.} In this simulation\footnote{The MATLAB code is available at: \url{http://bsp.brain.riken.jp/~zhougx/resources/mcode/demo_CIFE.zip}.} we generated a total of ten matrices $\mats{S}\in\Real^{5000\times10}$, $n=1,2,\ldots,10$, for which the first four columns were speech signals included in the ICALAB benchmark (named Speech4.mat) \cite{ICALAB}, and the other six components were drawn from independent standard normal distributions. The entries of the mixing matrices $\mats{M}\in\Real^{50\times10}$ were also drawn from independent standard normal distributions. We used the model $\mats{Y}=\mats{S}\mats{M}^T+\mats{E}$, where \mats{E} contains white Gaussian noise (SNR=20dB), and first employed the COBE, JIVE \cite{JIVE2013}, JBSS \cite{JBSSCCA}, and PCA methods\footnote{To achieve higher separation accuracy, we  used the \emph{princomp} function included in MATLAB to perform PCA. For COBE/COBE$C$, we simply used tSVD to perform dimensionality reduction, which has led to improved efficiency against our earlier version published at \url{http://arxiv.org/abs/1212.3913}. }
 to extract the common bases \mat{\bar{A}}, followed by the SOBI method  \cite{Belouchrani1997} to extract the latent common speech signals \mat{S} from \mat{\bar{A}}, see Section \ref{sec:LinkedBSS}. TABLE \ref{tabICA} shows the quantitative performance averaged over 50 Monte-Carlo runs, where separation accuracy SIR$_i$, i.e., the signal-to-interface ratio (SIR) of the $i$th estimated signal, was measured via
\begin{equation}
  \label{eqSIR}
  \text{SIR}(\mathbf{s},\widehat{\mathbf{s}})=10\log_{10}{\frac{\sum_t{s_t^2}}{\sum_t({s_t-\widehat{s}_t})^2}},
\end{equation}
where $s$ and $\widehat{s}$ are normalized random variables with zero mean and unit variance, and $\widehat s$ is an estimate of $s$. Observe that JIVE and COBE achieved higher SIRs than JBSS and PCA, although the performance of JBSS  improved after incorporating  SOBI. Moreover, although PCA has a close relation with COBE, the common features extracted by PCA are often contaminated by individual features; also while COBE and JIVE achieved almost the same separation accuracy, COBE was much faster (The time for dimensionality reduction has been included for all algorithms). Moreover, after we reduced the power of common components such that $\mats{\bar{s}}^T\mats{\bar{s}}=100$, $n=1,2,3,4$, COBE obtained similar accuracy whereas JIVE failed, which conforms with the analysis in Section 2.4. Also, the performance of JIVE was sensitive to the correct estimate of the rank of joint/common and individual components. If the ranks of individual components were mis-specified, for instance  as 7 (denoted as JIVE$^*$ in TABLE \ref{tabICA}), instead of the  actual 6, JIVE took more than 77 seconds to converge. Since the estimation of the number of components in \cite{JIVE2013} is quite time consuming and the performance is sensitive to selection of its parameters (for this instance, JIVE took more than two hours to estimate the rank); this limits the practical applications of JIVE for large-scale problems.

\begin{figure}
\centerline{
  \includegraphics[width=0.45\textwidth,height=0.25\textwidth]{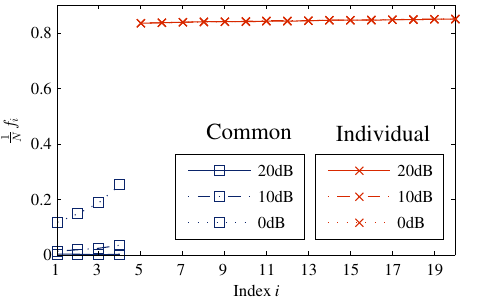}}
  \caption{Detection of the the number of common components by locating the GAP between the values of $\frac{1}{N}f_i$, under different noise levels.}
  \label{figf_i}
\end{figure}

\begin{figure}
\centerline{
  \includegraphics[width=0.48\textwidth,height=0.3\textwidth]{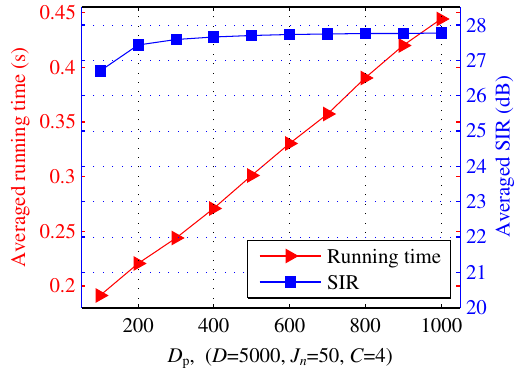}}
  \caption{Averaged performance of COBE after projecting the $D$-dimensional observations onto a lower $D_\text{P}$-dimensional space over 50 Monte-Carlo runs.}
  \label{figpcobe}
\end{figure}

The distinguishing properties of COBE verified by this simulation example include:
\begin{itemize}
  \item COBE is able to identify true common subspace even if the common components are relatively weak;
  
  \item Computational complexity of COBE depends only on the natural parameters: the size of the data and the number of common components $C$, making it much more efficient than JIVE;

  \item Finding the number of common components in COBE is physically intuitive and simple;

  \item The threshold $\epsilon$ within COBE also has physical interpretation---the degree of similarity between the components.
\end{itemize}

\figurename \ref{figf_i} illustrates the underlying principle of the estimation of the number of components by tracking the parameter $f_i$, while   \figurename \ref{figpcobe} illustrates the average (over 50 runs) performance in terms of the running time and separation accuracy of COBE with the observations projected onto a lower $D_{\text{P}}$-dimensional space by multiplying  with an $D_{\text{P}}\times D$ random matrix \mat{P}. As desired, the running time was almost linear in the dimension of the projected space $D_\text{p}$, illustrating that we may use projections to significantly improve the efficiency of COBE when the number of observations $D$ is very large.

\begin{table}[!th]
\caption{Performance comparison in linked BSS. The latent signals were estimated by applying the SOBI method to the common components extracted by each algorithm.}
\label{tabICA}
\begin{center}
\begin{tabular}{ l  c  c  c  c c }
\hline \hline
 Algorithm & $\mbox{SIR}_1$ & $ \mbox{SIR}_2$ & $\mbox{SIR}_3$ & $\mbox{SIR}_4$ & Runtime (s)\\
 \hline
COBE            & 21.1   &  24.3   & 27.2   & 24.6 &  0.1\\
COBE$C$          &  21.1   & 23.3   & 24.2   & 25.2 & 0.1 \\
JIVE               & 21.2   & 23.8   & 24.2   & 25.0 &  7.4 \\
JIVE$^*$             &  21.2   & 23.8   & 24.1   & 24.9 & 77.5 \\
JBSS            & 15.1   & 15.4    & 15.9   & 16.3  & 1.7 \\
PCA             & 15.8   & 17.1   & 17.8   & 19.4  & 0.5 \\
  \hline \hline
\end{tabular}
\end{center}
\end{table}

 \begin{figure}[!t]

    \centerline{
    \subfloat[The sources]{
    \includegraphics[width=0.24\textwidth,height=0.10\textwidth]{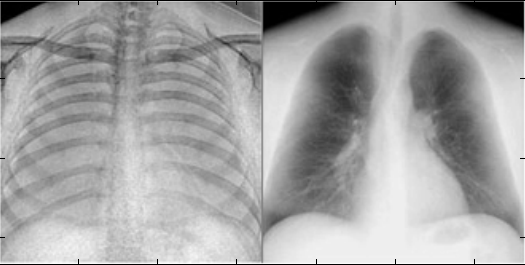}}
    \subfloat[The images extracted by CNFE]{
    \includegraphics[width=0.24\textwidth,height=0.10\textwidth]{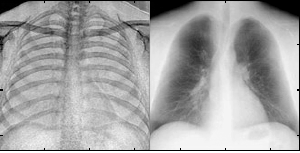}}
    }
    \centerline{
    \subfloat[Samples of the observations/mixtures] {
    \includegraphics[width=0.48\textwidth,height=0.11\textwidth]{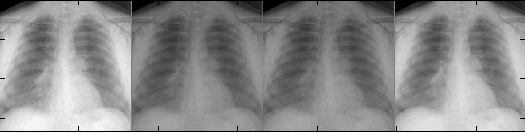}}
    }
    \centerline{
    \subfloat[Example images extracted by nLCA-IVM] {
    \includegraphics[width=0.48\textwidth,height=0.10\textwidth]{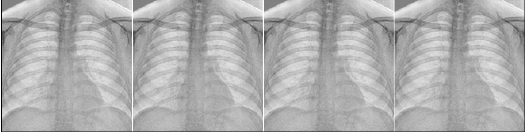}}
    }
    \caption{Illustration of common nonnegative feature extraction.}
    \label{figLung}
  \end{figure}

{\bf Dual-energy X-ray image decomposition.} Dual-energy chest X-ray imaging is a diagnostic tool for early signs of lung cancer \cite{nLCAIVM2010}, however, the ribs, clavicles overlapped with soft tissues, and environmental noise make it  challenging to detect affected lung nodules. To this end, we assumed a mixing model of bones, soft tissues and noise, where the former two were considered nonnegative common components. We considered four sets of sources $\mats{S}\in\Real^{26,896\times10}$, $n=1,2,3,4$, for which the first two common components were respectively the soft and bone tissues (both nonnegative) and the remaining eight components were drawn from independent uniform distributions between 0 and 1 to model the interference. The elements of nonnegative mixing matrices $\mats{M}\in\Real^{20\times 10}$, $n=1,2,3,4$, were also drawn from independent uniform distributions between 0 and 1. Then four set of observations were generated by using $\mats{Y}=\mats{SM}^T$. The sources in this example were highly correlated and could not be separated by  ICA methods, while the presence of random dense noise makes the separation by ordinary NMF on each single set of mixtures difficult. We applied COBE to extract a basis of common sources (soft tissues and bones) followed by CNFE to separate the soft tissues and bones. One typical realization is shown in \figurename \ref{figLung}(b), while \figurename \ref{figLung}(d) displays four samples of nonnegative components extracted by nLCA-IVM \cite{nLCAIVM2010}. Owing to dense noise, the identifiability conditions of nLCA-IVM were not satisfied, and nLCA-IVM could not extract the desired source images, while COBE performed nonnegative high correlation analysis well.

\begin{table*}
  \caption{Clustering Performance on Extended Yale B}
  \label{tabYale}
  \centering
  \begin{tabular}{ c | c c c c c  ||  c c c c c}
  \hline \hline
  \multirow{2}{*}{k}
  & \multicolumn{5}{c||}{Accuracy (\%)}   & \multicolumn{5}{c}{Normalized Mutual Information (\%)}   \\
  \cline{2-11}
  & PCA     & tSNE    & GNMF & MMCut & CIFE & PCA & tSNE & GNMF & MMCut & CIFE \\ \hline

10 & $43.9\pm{4.5}$    & $52.3\pm{6.3}$    & $60.3\pm{5.4}$    & $53.5\pm{3.4}$    & $\mathbf{71.2\pm{7.1}}$    & $38.5\pm{4.8}$    & $51.2\pm{5.8}$    & $61.6\pm{3.9}$    & $51.4\pm{2.3}$    & $\mathbf{67.3\pm{6.1}}$      \\
15 & $45.6\pm{6.2}$    & $48.7\pm{4.4}$    & $60.0\pm{6.4}$    & $56.0\pm{5.6}$    & $\mathbf{65.2\pm{6.1}}$    & $44.0\pm{5.8}$    & $51.1\pm{4.7}$    & $63.9\pm{5.5}$    & $55.9\pm{4.8}$    & $\mathbf{66.9\pm{5.2}}$      \\
20 & $41.3\pm{2.4}$    & $44.2\pm{2.5}$    & $52.4\pm{3.8}$    & $49.5\pm{4.0}$    & $\mathbf{62.2\pm{4.6}}$    & $43.5\pm{2.3}$    & $49.2\pm{2.8}$    & $59.6\pm{3.3}$    & $53.1\pm{3.3}$    & $\mathbf{65.7\pm{3.7}}$      \\
25 & $40.5\pm{2.3}$    & $40.2\pm{2.5}$    & $48.7\pm{2.7}$    & $46.5\pm{3.7}$    & $\mathbf{57.8\pm{2.9}}$    & $45.4\pm{2.1}$    & $46.2\pm{2.4}$    & $57.4\pm{2.5}$    & $51.9\pm{3.2}$    & $\mathbf{63.3\pm{2.3}}$      \\
28 & $37.3\pm{1.4}$    & $39.2\pm{1.5}$    & $47.0\pm{1.7}$    & $47.7\pm{1.4}$    & $\mathbf{57.1\pm{1.1}}$    & $8.2\pm{5.8}$    & $8.8\pm{6.3}$    & $10.8\pm{7.6}$    & $10.4\pm{7.4}$    & $\mathbf{12.0\pm{8.5}}$      \\
\hline
Avg. & 42.2    & 45.5    & 54.4    & 51.0    & {\bf63.3}    & 42.9    & 49.1    & 60.2    & 53.3    & {\bf65.5}     \\

\hline \hline
  \end{tabular}
\end{table*}

\begin{table*}
  \caption{Clustering Performance on PIE}
  \label{tabPie27}
  \centering
  \begin{tabular}{ c | c c c c c  ||  c c c c c}
  \hline \hline
  \multirow{2}{*}{k}
  & \multicolumn{5}{c||}{Accuracy (\%)}   & \multicolumn{5}{c}{Normalized Mutual Information (\%)}   \\
  \cline{2-11}
  & PCA     & tSNE    & GNMF & MMCut & CIFE & PCA & tSNE & GNMF & MMCut & CIFE \\ \hline

10 & $79.9\pm{0.5}$    & $34.6\pm{0.3}$    & $94.2\pm{0.8}$    & $63.7\pm{2.7}$    & $\mathbf{99.3\pm{0.2}}$    & $82.0\pm{0.3}$    & $39.8\pm{0.5}$    & $92.7\pm{0.7}$    & $71.7\pm{1.4}$    & $\mathbf{98.9\pm{0.3}}$      \\
20 & $86.8\pm{1.2}$    & $29.4\pm{0.7}$    & $90.5\pm{1.1}$    & $61.4\pm{0.5}$    & $\mathbf{100.0\pm{0.0}}$    & $88.3\pm{1.0}$    & $47.7\pm{0.2}$    & $92.8\pm{0.4}$    & $76.5\pm{0.9}$    & $\mathbf{100.0\pm{0.0}}$      \\
30 & $74.8\pm{0.0}$    & $40.8\pm{0.0}$    & $83.9\pm{0.0}$    & $60.2\pm{0.0}$    & $\mathbf{90.6\pm{0.0}}$    & $85.3\pm{0.0}$    & $60.2\pm{0.0}$    & $92.4\pm{0.0}$    & $77.0\pm{0.0}$    & $\mathbf{96.6\pm{0.0}}$      \\
40 & $67.8\pm{2.4}$    & $33.5\pm{0.9}$    & $74.7\pm{2.3}$    & $62.7\pm{0.2}$    & $\mathbf{86.3\pm{0.2}}$    & $83.6\pm{1.2}$    & $57.0\pm{1.0}$    & $88.5\pm{0.7}$    & $78.8\pm{0.2}$    & $\mathbf{95.5\pm{0.1}}$      \\
50 & $72.3\pm{0.7}$    & $36.4\pm{0.8}$    & $73.7\pm{0.6}$    & $60.4\pm{0.0}$    & $\mathbf{83.1\pm{0.2}}$    & $86.6\pm{0.2}$    & $61.1\pm{0.4}$    & $87.7\pm{0.3}$    & $79.1\pm{0.2}$    & $\mathbf{94.5\pm{0.1}}$      \\
60 & $76.3\pm{0.4}$    & $35.0\pm{0.6}$    & $75.4\pm{0.5}$    & $58.4\pm{0.3}$    & $\mathbf{84.3\pm{0.6}}$    & $87.2\pm{0.2}$    & $62.1\pm{0.3}$    & $88.4\pm{0.3}$    & $79.2\pm{0.0}$    & $\mathbf{94.6\pm{0.1}}$      \\
68 & $77.8\pm{0.5}$    & $30.2\pm{0.6}$    & $71.8\pm{1.1}$    & $60.3\pm{0.1}$    & $\mathbf{85.3\pm{0.1}}$    & $86.6\pm{0.2}$    & $60.6\pm{0.1}$    & $87.2\pm{0.3}$    & $80.2\pm{0.1}$    & $\mathbf{95.0\pm{0.0}}$      \\ \hline
Avg.  & 76.5    & 34.3    & 80.6    & 61.0    & 89.8    & 85.6    & 55.5    & 90.0    & 77.5    & {\bf96.4}     \\

  \hline \hline
  \end{tabular}
\end{table*}

{\bf Clustering Analysis Using Individual Features.} We considered two data sets:

{\emph{Extended Yale Database B\footnote{[Online]: \url{http://vision.ucsd.edu/~leekc/ExtYaleDatabase/ExtYaleB.html.}}.} The database contains 16,128 images of 28 human subjects under 9 poses and 64 illumination conditions \cite{YaleB}. 
A total of 5,820 approximately frontal faces were automatically detected by using  the classification and regression tree analysis (CART) and cropped  for our clustering analysis.

{\emph{PIE Database\footnote{[Online]: \url{http://vasc.ri.cmu.edu/idb/html/face/.}}.}} This database is a collection of face images of 68 persons taken under different poses, illumination conditions, and expressions. We used a pre-processed version from \cite{GNMF2011PAMI} which consists of 2,856 full frontal face gray scale images taken at the pose c27.

All images were re-scaled to the size of $32\times32$. We randomly selected $K$ clusters each time, and repeated the experiment 50 times for each selected $K$. In each run, the images were first  permuted randomly and then split into $N=\lfloor\frac{T}{50}\rfloor$ groups to form multi-block data \mats[n]{Y} with $J_n\approx 50$, $n=1,2,\ldots,N$, and $T$ is the number of faces, (each group consisted of face images from unknown different clusters). COBE was used to extract the common features followed by CNFE to obtain nonnegative common features. The number of common components was specified as 2 in all experiments, and the two t-SNE components of their individual parts \mats{\breve{Y}} were used to cluster the data by using $K$-means (see \cite{tSNE2008} for the t-SNE method). As $K$-means is influenced by initial centers of clusters, we replicated $K$-means 20 times in each run. Two widely used performance indices: Accuracy (\%) and Normalized Mutual Information (NMI) were adopted to evaluate the clustering results, see \cite{GNMF2011PAMI} for detailed definitions of these two metrics. The proposed method was compared with PCA with $K$ principal components, GNMF \cite{GNMF2011PAMI} (using their recommended settings), and the improved MinMax Cut (MMCut) method \cite{MMCut2010}. Except for  MMCut that  returns the cluster indicators directly, all the other methods used the $K$-means to cluster the  features extracted by them with the same configuration. To illustrate that the performance of the proposed method was not completely due to t-SNE, two t-SNE components of the original  data were also used as features for clustering. The clustering performance of these algorithms is detailed in TABLE \ref{tabYale} and \ref{tabPie27}, respectively, showing that after removing the common features pertaining to all samples, the clustering performance were significantly improved.

In \figurename \ref{figParac} we next illustrate how the number of common components, $C$, influenced the clustering performance , where the values of $C$ varied from 1 to 15. For the PIE dataset we used the first 40 categories while for the Extended Yale database B we used the first 20 categories. From the figure, for both datasets the clustering performance was significantly improved after removing the first 2-3 common components. While how to set the optimal parameter $C$ blindly still remains challenging for practical applications; empirical results suggest that performance degradation degeneration caused by the overestimation of $C$ is not significant if our interest is the individual features. To show this, consider
\begin{equation}
\begin{split}
  \mats{\breve{Y}}=& \mats{Y}-\sum\nolimits_{c=1}^C \mat{\bar{a}}_c \mats[n,c]{\bar{b}}^T  \\
                            = & \mats{Y}-\sum\nolimits_{c=1}^C \mat{\bar{a}}_c (\mat{\bar{a}}_c^T\mats{Y}).
\end{split}
\end{equation}
where $\mat{\bar{a}}_c$ and $\mats[n,c]{\bar{b}}$ are the $c$th column of \mat{\bar{A}} and \mats{\bar{B}}, respectively. For an overestimated $C$ the correlations between $\mat{\bar{a}}_C$ and the components in \mats{Y} are quite small (close to 0, ideally), and $\mats[n,C]{\bar{b}}=\mat{\bar{a}}_C^T\mats{Y}$, i.e., the projection of \mats{Y} on $\mat{\bar{a}}_C$ becomes very small. In other words, the loss of individual features tends to be very small if $C$ has been slightly overestimated.

 \begin{figure}[!t]
 \centerline{
    \includegraphics[width=0.49\textwidth]{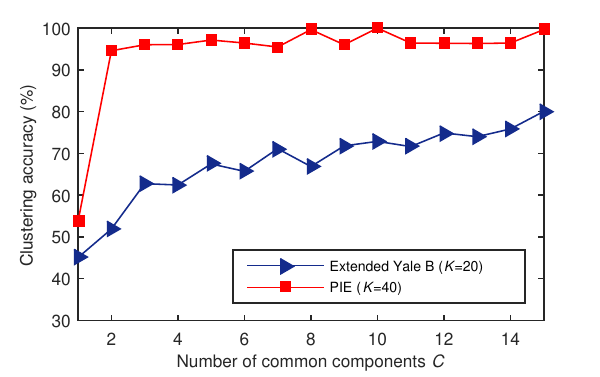}}
    \caption{Influence of the number of common components $C$ on the clustering performance.}
    \label{figParac}
  \end{figure}



 \begin{figure}[!t]

    \centerline{
    \subfloat[The eight categories of the ETH-80 database]{
    \includegraphics[width=0.5\textwidth]{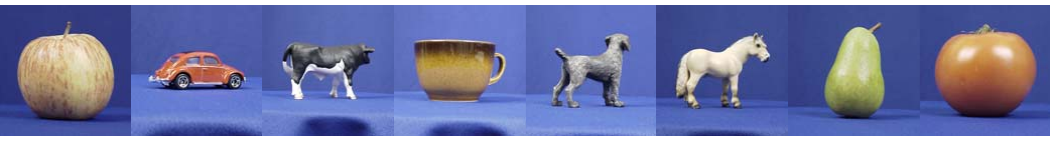}}
    }
    \centerline{
    \subfloat[The 10 objects in the forth category]{
    \includegraphics[width=0.5\textwidth]{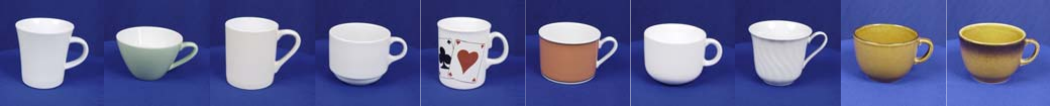}}
    }
    \caption{The ETH-80 database.}
    \label{figETH80}
  \end{figure}

 \begin{figure}[!t]
    \centerline{    \includegraphics[width=0.5\textwidth]{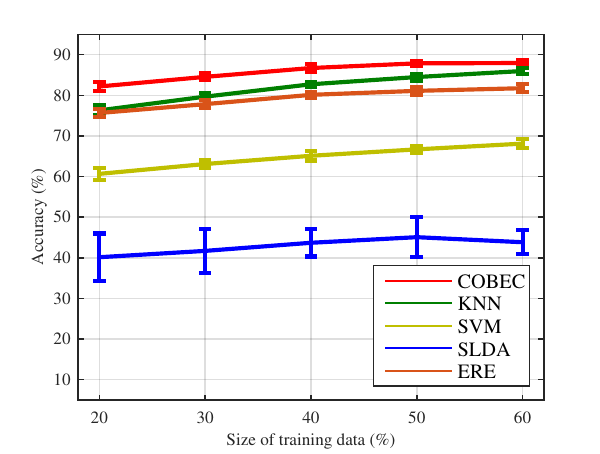}    }
    \caption{Mean values and standard derivations for the accuracy in the classification of the ETH-80 database over 20 random runs.}
    \label{figClassify}
  \end{figure}

{\bf Applications in classification.} The classification performance of the proposed method  was evaluated by using the ETH-80 Dataset.
 The ETH-80 dataset consists of a total of 3,280 images grouped in 8 categories containing 10 objects with 41 views per object, spaced equally over the viewing hemisphere \cite{ETH80}. Each category contains 10 different objects. Although the objects belonging to the same category  share some common features, they also have their individual features different from the other objects in the same category, which makes this database widely adopted to evaluate classifying methods. See \figurename \ref{figETH80} for the 8 categories in ETH-80 and the 10 objects in the forth category.
We compared our CIFE based classifier (see \figurename \ref{figCFEClassify}) with the KNN classifier included in MATLAB 2010b, the SVM classifier \cite{libsvm}, the Eigenfeature Regularization and Extraction method \cite{ERE2008} (ERE), and the shrinkage LDA \cite{ShrinkageLDA} (SLDA). As the ERE method needs to perform eigenvalue decomposition of full covariance matrices, we used the grayscale images with the size of $32\times 32$ to avoid ERE running out of memory. For the KNN classifier 1 nearest (measured by correlation) neighbor was used for classification. For SVM, we used 5-fold cross validation and the best parameters of $c$ and $\gamma$ were found using grid search following the guidelines provided by the authors. While the SLDA can determine the optimal parameters automatically, the ERE has two tunable parameters: $\mu$ that is used to determine the noise region and the number of features $d$.  For fair comparison we set $\mu=1$, as suggested by the authors. To achieve the best performance $d$ was selected as the number of nonzero eigenvalues. In the CIFE based classifier,  we split the training samples belonging to each class into $\max(2,\lfloor{T}/{50}\rfloor)$ subgroups and then used COBE$C$ to extract their common features, where $T$ was the number of training samples in this class. For simplicity, we empirically set $C=\min_n J_n\times80\%$, although the optimal $C$ and number of subgroups could be estimated by using cross validations.  Correlation was adopted as the matching score to classify a new test sample (see \eqref{eqMatchingScore} and \figurename \ref{figCFEClassify} for details). In each run, we randomly selected a certain percentage of samples as the training data and the remainder as the test data. The mean values and standard derivations of classification accuracy over 20 random runs are plotted in \figurename \ref{figClassify}, showing that the CIFE based classifier yielded the best classification accuracy among the classifiers considered.

\section{Conclusions and Future Work}
A new scheme for common and individual feature extraction (CIFE) for naturally linked multi-block data has been proposed, together with two new efficient algorithms for the extraction of common orthogonal bases (COBE) according to whether the number of common components is known or not. We have also introduced the concept of linked Blind Source Separation (BSS) of multi-block data in order to perform effective task-dependent feature extraction in common and individual subspaces rather than in a global high dimensional space. The proposed CIFE scheme  has been validated on classification and clustering tasks, by exploiting the separated common and individual features. Comprehensive simulations have illustrated the ability of the proposed methods to extract common features existing  in multi-block data efficiently and accurately. 

In this study we have concentrated on developing a unifying and versatile scheme of common and individual feature extraction.  Questions that remain to be investigated in the  future include:
\begin{enumerate}
  \item The number of common features, i.e., $C$, which is controlled by the parameter $\epsilon$, often plays a quite important role in practical applications: it controls the degrees of similarity (correlation, in this paper) of  common components extracted from different data sets.  However, determining its optimal theoretical value is challenging and needs further investigation;

  \item For some practical applications, we need to split the data into subgroups manually in order to discover their common features. Optimal grouping of data is a important factor to achieve better performance;

  \item In this paper we considered common features in one dimension. Further extensions of the proposed method to higher-order data will yield general and flexible CIFE tools for tensor data. 
  
  \item We provided two example applications of CIFE and corresponding experimental evidences to justify their validity and high performance. However, how to unlock the full potential of CIFE in machine learning deserves further investigation. 
\end{enumerate}

\bibliographystyle{IEEEtran}

\end{document}